# A Learning-from-Observation Framework: One-Shot Robot Teaching for Grasp-Manipulation-Release Household Operations

Naoki Wake, Riku Arakawa, Iori Yanokura, Takuya Kiyokawa, Kazuhiro Sasabuchi, Jun Takamatsu, and Katsushi Ikeuchi

*Abstract*— A household robot is expected to perform various manipulative operations with an understanding of the purpose of the task. To this end, a desirable robotic application should provide an on-site robot teaching framework for non-experts. Here we propose a Learning-from-Observation (LfO) framework for grasp-manipulation-release class household operations (GMR-operations). The framework maps human demonstrations to predefined task models through one-shot teaching. Each task model contains both high-level knowledge regarding the geometric constraints and low-level knowledge related to human postures. The key idea is to design a task model that 1) covers various GMR-operations and 2) includes human postures to achieve tasks. We verify the applicability of our framework by testing an operational LfO system with a real robot. In addition, we quantify the coverage of the task model by analyzing online videos of household operations. In the context of one-shot robot teaching, the contribution of this study is a framework that 1) covers various GMR-operations and 2) mimics human postures during the operations.

## I. INTRODUCTION

With an aging population, the introduction of robots into a household environment becomes desirable to compensate for the reduced physical labor able to be completed by the elderly. A survey of older adults revealed physical tasks that are preferable to be replaced by robots [1]. One type of the tasks is the manipulation of objects regardless of purpose or context. For example, opening and closing doors or drawers, reaching for objects, fetching objects, and picking up heavy objects. We refer to the type of tasks as grasp-manipulation-release class operations (hereafter, GMR-operations). A household robot is expected to perform various GMR-operations.

In order to program robots to perform GMR-operations, human helpers can teach them to perform tasks at home to meet the purpose of the user. However, such helpers are typically non-experts in robot programming. In an ideal framework, robots that can perform some basic GMR-operations would be optimized in each home through on-site instruction by non-experts. The key system requirements of such a robot teaching framework are as follows: 1) teaching should be easy and preferably completed via one-shot teaching, 2) teachable operations should cover basic GMR-operations, and 3) the robot should be able to mimic human postures, which contain implicit information required to achieve a task with a specific purpose [2].

One framework able to meet these system requirements is Learning-from-Observation (LfO) (Fig. 1). LfO aims to teach robots a sequence of what-to-do and how-to-do instructions through a one-shot human demonstration [3]. What-to-do and how-to-do are referred to as a task and skill parameters, respectively, and a pair of task and skill parameters is referred to as a task model. An advantage of LfO is that it can cover various operations within the scope of a predefined task model.

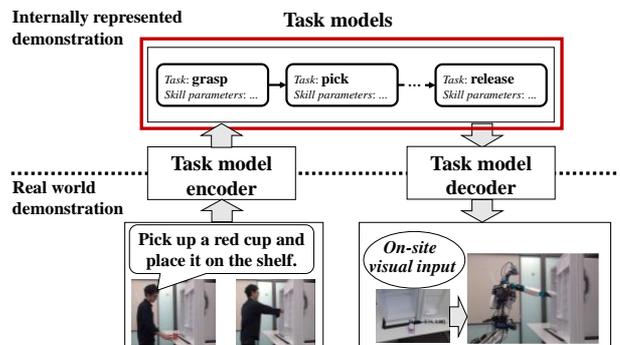

Figure 1. Learning-from-Observation framework. The red rectangle depicts the main research object of this paper.

Another popular robot teaching framework is one-shot Learning-from-Demonstration (LfD) [4], [5]. Although LfO and LfD are similar in that they map human demonstrations to robot movements, they are based on a different philosophy. While LfO is based on the task-oriented programming approach, which uses pre-defined intermediate task representations [3], LfD is based on the machine-learning approach, which aims to obtain intermediate task representations through repeated observation of human demonstrations [6]–[8].

Although research on LfO and LfD has shown the importance of intermediate task representations for one-shot robot teaching, the applications are still limited to specific domains, such as part assembly, [3] knot tying [9], block building [10], rotating [11], or scooping [12]. For teaching robots various GMR-operations, the key factor appears to be the design of the intermediate task representation, which covers basic GMR-operations.

In this study, we design a task model for GMR-operations based on the LfO philosophy. Specifically, we target door or drawer opening and pick-carry-place tasks as basic GMR-operations. To meet the system requirements, we design the task model to contain 1) a task set that covers state transitions in the basic GMR-operations, 2) traditional skill parameters in terms of LfO, and 3) additional skill parameters to represent human posture. To verify the applicability of the proposed task model, we test an operational LfO system with the basic GMR-operations. An ideal task set should cover not only the basic GMR-operations, but also as many household GMR-operations as possible. Thus, we evaluate the coverage of household GMR-operations by analyzing online instruction videos. In the context of one-shot robot teaching, the contribution of this study is a robot programming framework that 1) covers various GMR-operations and 2) mimics human postures during the operations.

## II. RELATED WORKS

As described in the Introduction, we propose a robot teaching framework that 1) enables one-shot teaching, 2) covers basic GMR-operations, and 3) mimics human postures. In this section, we describe previous robot teaching frameworks in the context of these system requirements.

### A. Learning-from-Observation (LfO)

LfO aims to map a one-shot demonstration to robot actions via well-designed intermediate task representations (i.e., task models) [3]. Because of its quick and intuitive manner of teaching, LfO is a suitable robot teaching framework for non-experts. However, previous LfO applications have been limited to specific domains, such as part assembly, knot tying, grasping, and dancing [3], [9], [13]–[19]. These limitations may be due to difficulties designing a task model that supports generic operations such as GMR-operations.

Considering the system requirements, the desired task model for GMR-operations should include: 1) a task set that covers GMR-operations and 2) skill parameters that support human postures. To the best of our knowledge, a task set suitable for GMR-operations has not yet been proposed. Although human posture has been considered in LfO for gestures [13], it has not been considered for GMR-operations. Our proposed task model extends LfO to include human posture information and cover household GMR-operations beyond a specific domain.

### B. Learning-from-Demonstration (LfD)

LfD or Programming-by-Demonstration (PbD) is another popular robot teaching framework. In this paper, we use LfD to refer to both PbD and LfD. Previous LfD research has involved high-quality surveys [6]–[8], with the majority aiming to obtain state-action pairs through repeated demonstrations. However, repeated teaching is not preferable for rapid robot teaching, and recent studies have proposed one-shot LfD for faster robot teaching [4], [5], [10], [11], [20], [21]. Although one-shot teaching has the advantage of simplicity of use [4], [22], its applications have previously been limited to specific domains such as block building, scooping, or rotating [10]–[12], potentially because it is difficult to adapt knowledge derived from a dataset to a novel demonstration [4], [10], [12], [21], [22]. This challenge is referred to as the domain-adaptation problem.

To address this problem, more recent research has focused on employing other learning frameworks, such as reinforcement learning (RL) [23]–[25], general adversarial network [26], and meta-learning [27]–[29]. However, these studies have not yet been applied to a real robotic system [24], [29] or were limited to a specific domain such as locomotion, scooping, and pick-and-place tasks [23], [26]–[28]. To extend these approaches for GMR-operations, a large dataset is required to enable domain adaptation. Moreover, these approaches have neglected human postures as models for generating robot postures, which may be due to the difficulty inherent in designing a way to evaluate "human-likeness." For a case in which the desired state is difficult to design explicitly, our simple approach, which directly encodes on-site human postures, could solve the problem of how to mimic human postures.

## III. DESIGN OF TASK MODELS

The key idea of this study is to design a task model that 1) covers various GMR-operations and 2) includes human postures to achieve tasks. This section explains the design of the task set, the encoding of human postures as a skill parameter, and the design of other skill parameters corresponding to the geometric constraints involved in a task.

### A. Definition of Tasks

In LfO, a task is defined as a transition of a target object's state. An example is a contact state between polyhedral objects for part assembly [3] or a topology of a string for knot tying [9]. In this study, we defined a state as a contact state between a target object and an environment (Fig. 2(a)). As the scope of this study is manipulative operations, we referred to a manipulation motion taxonomy in household operations [30]. The literature has defined a set of motion types from a robotics perspective, considering both contact and non-contact states. Furthermore, a prismatic and revolute motion trajectory has been considered in the contact state. In this study, we included terminal states of the trajectories. As a consequence, we defined the states as non-contact (NC), planar contact (PC), prismatic contact (PR), one-way prismatic contact (OP), revolute contact (RV), or one-way revolute contact (OR). Considering that a target object may include mechanical linkages in several manipulation tasks (e.g., opening a door) [1], we additionally checked that those states were consistent with Mason's definition of basic states for mechanical linkages [31], [32]. Fig. 2(b) shows the possible state transitions (i.e., task set) between the states. We also included grasp and release in the task set because these actions are accompanied by a transition in the contact state between a robot end effector and a target object.

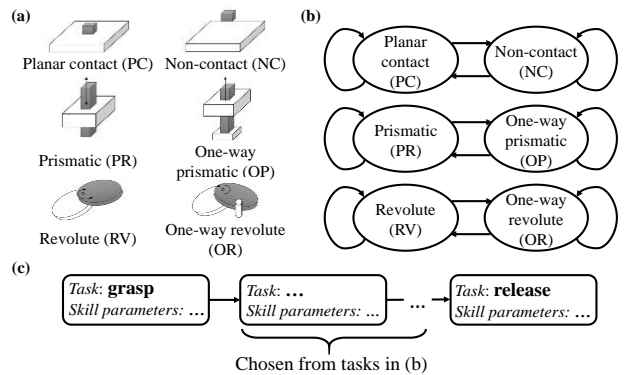

Figure 2. Representation of a GMR-operation. (a) The six contact states defined in this study, where the grey object is a target object. (b) Task sets defined by possible state transitions. (c) Definition of a GMR-operation using a sequence of task models. The tasks are chosen from the task set defined in (b).

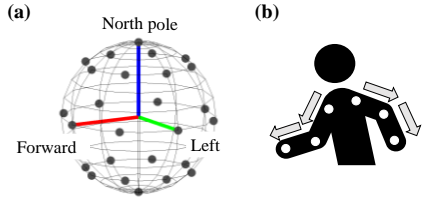

Figure 3. Digitization of human upper-body postures. (a) 26-point directions on the unit sphere. The forward direction is defined as the axis orthogonal to a surface including the spine and shoulders. (b) Joints included to define arm postures. The relative directions between joints, which are depicted as gray arrows, are mapped to the directions in (a).

Fig. 2(c) shows the definition of a GMR-operation using task models. A GMR-operation starts with a grasp and ends with a release, with a sequence of tasks in between. For example, when a human carries an object from one location to another, the sequence of tasks is *grasp-object*, *pick-up-object (PC-NC)*, *carry-object (NC-NC)*, *place-object (NC-PC)*, and *release-object*. As another example, when a human opens a fridge, the sequence of tasks is *grasp-door*, *open-door (OR-RV)*, *rotate-door (RV-RV)*, *stop-opening (RV-OR)*, and *release-door*.

### B. Skill Parameter of Human Posture

We encoded a human posture into spatially digitized 26-point directions on the unit sphere (Fig. 3(a)). The idea is based on existing human motion representations [13], [33]. Digitization has several advantages over using raw data; it reduces sensor noise in raw data and allows filtering of obvious detection errors in human posture. For example, unnaturally twisted arm poses can be checked against a table that defines possible human postures. Human postures during a demonstration are encoded frame-by-frame then stored as skill parameters.

In this study, we focused on arm postures because each human body part plays different roles during a reaching motion; the human arm has control over the reaching motion and the human trunk adds additional length when reaching [34]. Considering that grasping or releasing is a form of reaching for a position, we assumed that the arm postures play a crucial role in a GMR-operation. Therefore, an arm posture was represented as a combination of the four directions of the upper arm joints (i.e., forearm and lower arm, on each side; see Fig. 3(b)).

### C. Skill Parameters of Geometric Constraints

We categorized the tasks into three classes in terms of robot control [2]:

- Position goal task: to achieve a desired state by applying a positional shift $p$ to a target object (i.e., NC-NC).
- Force goal task: to achieve a desired state by applying force $f$ to a target object (e.g., NC-PC).

TABLE I. SKILL PARAMETERS OF MANIPULATION TASKS.

| Tasks | Position parameters | Force parameters |
|---|---|---|
| NC-NC | Waypoints | - |
| NC-PC, OP-PR, OR-RV | - | Detaching axis direction; force on the axis |
| PC-NC, PR-OP, RV-OR | - | Attaching axis direction; force on the axis |
| PC-PC | Trajectory on maintaining dimension (2D plane) | Surface normal axis direction; force on the axis |
| PR-PR | Trajectory on maintaining dimension (1D distance) | Plane orthogonal to trajectory; force in the plane dimension |
| RV-RV | Trajectory on maintaining dimension (angle; radius) | Axis direction to rotation center; force on the axis |

TABLE II. SKILL PARAMETERS OF THE GRASP AND RELEASE TASKS.

| Tasks | Parameters filled by daemon | Parameters filled on-site |
|---|---|---|
| grasp | Object name, object attribute, grasp type, manipulating hand, grasp location | Grasp position |
| release | Release location | Release position |

- Hybrid goal task: to achieve a desired state by applying a positional shift $p$ and force $f$ to a target object (e.g., PC-PC).

In order to achieve these tasks, two types of skill parameters were required: 1) position parameters to apply the positional shift $p$ and 2) force parameters to apply the force $f$. Table 1 illustrates the skill parameters of each task. The position parameter was defined as the trajectory to translate a target object. The force parameter was defined as a force vector, in which a force control was required. Note that the force vector was not to cause an object translation, but to control the contact between a target object and an environment.

Table 2 illustrates the skill parameters of the grasp and release task. We divided the skill parameters into two classes: those obtained through a demonstration and those obtained at the time of robot execution. For example, a grasp position was calculated at the moment when the robot executed the task because a target-object position is not invariant.

## IV. IMPLEMENTATION OF TASK-MODEL ENCODER

This section briefly explains the pipeline implemented to substantiate the task models (Fig. 4). The purpose of this section is to explain the feasibility of obtaining the designed skill parameters from human demonstrations. A detailed explanation of the implementation falls outside the scope of this paper and is described in [35].

We assumed that a human demonstrates a GMR-operation with a co-occurring instruction (e.g., "pick up a red cup on the table, like this.") and that tasks shall be associated with verbs by a pre-defined knowledge database. Thus, a demonstration was fed into the pipeline as verbal and visual inputs. The verbal input was verbal instructions transcribed using a cloud speech recognition service, and the visual input was a time-series of RGB-D images and a time-series of human-skeleton poses obtained by a vision sensor [36]. The pipeline outputted substantiated task models, which contained recognized task and skill parameters (Fig. 2(c)). The pipeline consisted of three modules: 1) the grasp and release detector, 2) the task detector, and 3) the skill parameter finder.

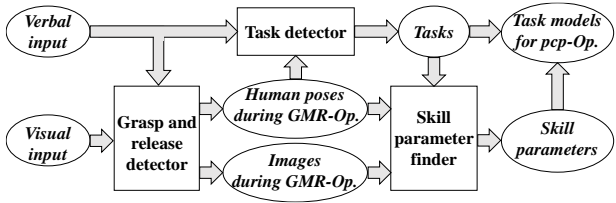

Figure 4. Pipeline used to substantiate the task model.

### A. Grasp and Release Detector

This module analyzed the verbal and visual inputs to segment the visual input at times when a grasp and a release occurred. To this end, we developed a verbal-based Focus-of-Attention (FoA) system [35]. In brief, the module analyzed the distance between a hand and a target object extracted from the verbal input. The segmented visual input was analyzed in the following modules.

### B. Task Detector

The detector recognized demonstrated tasks by analyzing the verbal input and the segmented skeleton poses. A task was determined by referring to a knowledge database that associated verbs with task candidates (Table 3). For example, the verb "open" was associated with two candidates: OP-PR-PR and OR-RV-RV. When multiple candidates existed, one was selected by analyzing the shape of the trajectory of the manipulating hand. The module serialized the verbs found in the database into a task sequence in order of their utterances.

### C. Skill Parameter Finder

At this point, the pipeline had obtained the recognized tasks and the visual input that was segmented at times when a grasp and a release occurred. This module obtained the skill parameters for the tasks by analyzing the segmented visual input. To this end, several daemon processes were invoked.

#### 1) Position Parameters

A daemon analyzed the trajectory of a manipulating hand to obtain the position parameters. For an NC-NC task, the position parameter was obtained as a spatially discretized manipulating-hand trajectory (i.e., waypoints). For PC-PC, PR-PR, and RV-RV tasks, the position parameter was obtained as the parameters of 2D plane fitting, line fitting, and circle fitting, respectively, of the manipulating-hand trajectory.

#### 2) Force Parameters

The force parameters consisted of the direction of the force applied and the magnitude of the force. In the case of the force goal task (e.g., NC-PC), the direction of the force can be identified by analyzing the acceleration of human hands. In the case of the hybrid goal task (e.g., PC-PC), the direction of the force was identified as the axes orthogonal to the space where the positional shift was allowed. In both cases, the magnitude of the force was filled by a default value that can be modified in a robot execution module if necessary.

#### 3) Skill Parameters for the Grasp and Release Task

The verbal-based FoA system [35] outputted several skill parameters for the grasp and release task, such as target object name, object attributes, and manipulating hand laterality. Thus, daemons obtained the remaining skill parameters: grasp type, grasp location, and release location.

A grasp type is appropriately selected by a demonstrator according to the purpose of the task. For example, in the case of placing a cup on a shelf with a narrow space above and below, it is reasonable to grasp the side surface of the cup. On the other hand, in the case of placing a cup on top of a tray of other cups, it is reasonable to grasp the top surface of the cup. A daemon recognized one of human grasp types [37] using a pipeline that leveraged an object affordance [38].

The grasp and release locations were defined as locations where the grasp and release occurred in an environment model. The location was obtained as a label of a semantically segmented 3D area, such as an "above-a-shelf area," by matching the model with the positions of the manipulating hand when the grasp and release occurred. At the time of robot execution, the task-model decoder calculated the grasp and release positions inside the locations.

## V. EXPERIMENTS

### A. One-shot Teaching by an Implemented LfO System

We tested an operational LfO system to verify the applicability of the proposed task model. We used a humanoid robot, Seednoid [39], as the LfO agent because the robot has a pair of 7-DOF arms as well as a movable waist to enable various manipulations. This section demonstrates two representative cases: "pick-carry-place a cup" (PC-NC-NC-PC) and "open a fridge" (OR-RV-RV). We chose these cases because we targeted door or drawer opening and pick-carry-place as the basic GMR-operations and because revolute motion is a general description of linear motion, including prismatic motion. Detailed explanations about the encoding and execution of the task models are described in [35] and [2], respectively.

Fig. 5 shows the results of teaching the "pick-carry-place a cup" operation. For the teaching, a demonstrator picked up a red cup with the right hand and placed the cup on a shelf with the following verbal instruction: "Pick up a red cup and place it on the shelf." Fig. 5(a) shows part of the substantiated task models. The robot successfully performed the demonstrated operation by decoding the task model, suggesting the applicability of the task model (Fig. 5(b)).

Fig. 6 shows the results of teaching the "open a fridge" operation. For the teaching, a demonstrator opened a fridge door with the right hand with the following verbal instruction: "Open the fridge." Fig. 6(a) shows part of the substantiated task models. The robot successfully performed the demonstrated operation by decoding the task model, suggesting the applicability of the task model (Fig. 6(b)).

### B. Evaluation of the Coverage of Household Operations

When considering the adequacy of the task model, the task set should cover not only the basic GMR-operations, but also as many household GMR-operations as possible. That is, a wide range of household GMR-operations should be represented by the tasks (i.e., state transitions) shown in Fig. 2(b). Actions are generally closely related to verbs. Therefore, we examined the association between state transitions and action verbs, which are spoken in YouTube instruction videos about daily chores. The categories of daily chores selected in this study were "floor cleaning," "carpet cleaning," "furniture cleaning," and "cooking." We focused on these categories because they are representative household operations in the field of home economics [40] and because cooking and cleaning are included in the tasks that older people want a robot to perform [1]. Twenty videos for each category were chosen by searching for videos with a query of "how to do *," where * was replaced by a category name.

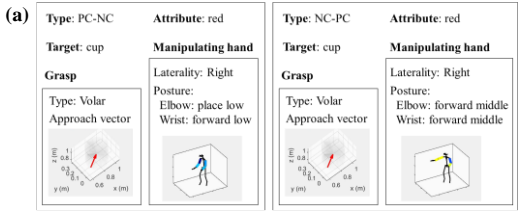
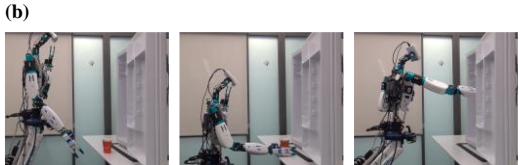

Figure 5. Results of LfO for "pick-carry-place a cup." (a) Substantiated task models with representative skill parameters. (b) Robot execution.

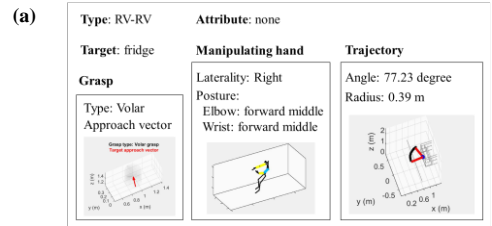
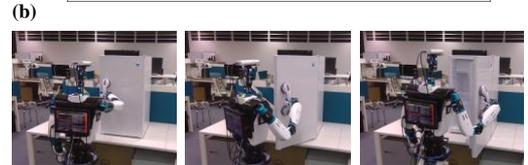

Figure 6. Results of LfO for "open a fridge." (a) A substantiated task model with representative skill parameters. (b) Robot execution.

We extracted the action verbs using the following procedures. First, all verbs were extracted using the Stanford parser, which is a widely used language parser in the field of natural language processing [41]. Next, to reduce the number of analyzed verbs, we chose verbs in the top 100 appearance probability list for each category. By dropping duplicates across categories, we extracted 239 unique verbs. Finally, we excluded verbs unrelated to a manipulation action such as "see," "go," "leave." The remaining 51 verbs were investigated (Table 4).

We then annotated a possible sequence of state transitions onto the 51 action verbs (Table 3). Interestingly, all the verbs were mapped to at least one sequence of state transitions defined in Fig. 2(b). This result suggests that the task set reasonably covered various household GMR-operations associated with verbal instructions.

## VI. DISCUSSION

Although many robot teaching frameworks have been proposed, their applications have so far been limited to specific domains such as locomotion, scooping, and pick-and-place [23], [26]–[28]. Therefore, to create a robot application for teaching GMR-operations, we extended an existing one-shot teaching framework, LfO. The key objectives were 1) to design a task model that covers the basic GMR-operations and 2) to include human posture as a skill parameter. In the context of one-shot robot teaching, our research highlights not only the effectiveness of intermediate task representation, but also the importance of skill parameters that are based on analyses of human behavior [30], [34], [37].

Experiments with a real humanoid robot successfully taught human posture and task constraints for several basic GMR-operations. In addition, the proposed task set was demonstrated to cover several household GMR-operations that were verbally taught in online instruction videos. The results suggest the applicability of the proposed framework beyond the basic GMR-operations. We are aware that the scope of this study may cover only part of the whole household operations. Nevertheless, we believe that assisting with even the basic GMR-operations will greatly contribute to the independent living of older adults, as suggested by a previous study [1].

The aim of this study was to design task models for GMR-operations. Robust encoding and decoding of the task models fell out of the paper's scope. For a robust LfO system, bottom-up learning methods should be employed both for the encoding and decoding. In particular, skill refinement using RL has long been a robot manipulation topic of intense interest [42]–[44]. In further research, we plan on employing RL frameworks using an algorithm to derive reward functions from substantiated task models.

TABLE III. ASSOCIATION BETWEEN VERBS AND TASK SEQUENCES.

| Representative verbs | Candidate of task sequences |
|---|---|
| take, remove, pick, lift, raise | PC-NC-NC |
| put, click, lay, restore, weigh, chop, place, slice, cut, press | NC-NC-PC |
| mop, burnish, wipe, polish, vacuum, scratch, buff, sweep, rub, scrub, dust, paint, scrape | PC-PC |
| dip, soak, carry, spray, mix, pour, stir, shake, tilt, hold | NC-NC |
| move, bring, flip, fold | PC-NC-NC-PC |
| pull, twist, turn, plug, open | OP-PR-PR or OR-RV-RV |
| cover | PR-PR-OP or RV-RV-OR |
| drop, fall, grab | grasp or release |

TABLE IV. SPECIFICATIONS OF THE ANALYZED VIDEOS.

|  | Floor | Carpet | Furniture | Cooking |
|---|---|---|---|---|
| Total length of the video (s) | 4764 | 10562 | 4783 | 8808 |
| Num. of unique verbs (count) | 346 (2204) | 425 (13459) | 308 (2524) | 268 (1843) |
| Num. of analyzed verbs (count) | 124 (1926) | 100 (12314) | 113 (2279) | 111 (1644) |
| Appearance rate of the analyzed verbs | 0.87 | 0.91 | 0.90 | 0.89 |